\documentclass[runningheads]{llncs}
\usepackage{graphicx}
\usepackage{subfigure}
\usepackage{multirow}
\usepackage{algorithmic}
\usepackage{amsmath,amssymb,amsfonts}
\usepackage{url}
\usepackage{hyperref}
\begin{document}
\title{MS-KD: Multi-Organ Segmentation with \\Multiple Binary-Labeled Datasets}
%
%

\author{
Shixiang Feng 
\and
Yuhang Zhou 
\and
Xiaoman Zhang 
\and
Ya Zhang
\and
Yanfeng Wang
}
\authorrunning{Y. Zhou et al.}
\institute{Cooperative Medianet Innovation Center, Shanghai Jiao Tong University 
\email{\{fengshixiang, zhouyuhang, xm99sjtu, ya\_zhang,wangyanfeng\}@sjtu.edu.cn}}

\maketitle              
 \begin{abstract}
Annotating multiple organs in 3D medical images is time-consuming and costly. Meanwhile, there exist many single-organ datasets with one specific organ annotated. This paper investigates how to learn a multi-organ segmentation model leveraging a set of binary-labeled datasets. A novel Multi-teacher Single-student Knowledge Distillation (MS-KD) framework is proposed, where the teacher models are pre-trained single-organ segmentation networks, and the student model is a multi-organ segmentation network. 
Considering that each teacher focuses on different organs, a region-based supervision method, consisting of logits-wise supervision and feature-wise supervision, is proposed. Each teacher supervises the student in two regions, the organ region where the teacher is considered as an expert and the back-ground region where all teachers agrees. 
Extensive experiments on three public single-organ datasets and a multi-organ dataset have demonstrated the effectiveness of the proposed MS-KD framework.

\end{abstract}

\section{Introduction}
Multi-organ segmentation is an important step towards many clinical applications such as disease diagnosis. 
However, due to the considerable annotating cost for 3D medical images segmentation, most benchmark datasets only provide segmentation mask of a single organ.
To address this problem,
several recent studies propose to leverage multiple single organ datasets to train a multi-organ segmentation model. Condition-based methods \cite{dmitriev2019learning,zhang2020dodnet} encode each segmentation task as task-aware prior to guide the network to segment the task-related organ. Several ways of conditioning the network are proposed, such as incorporating organ class information into the intermediate activation signal~\cite{dmitriev2019learning}, or incorporating the task-aware prior to generate dynamic convolution filters with a dynamic on-demand network (DoDNet)~\cite{zhang2020dodnet}.
However, the above condition-based methods require several rounds of inference to get the final segmentation, which is computationally inefficient. 
Pseudo-label based methods \cite{zhou2019prior,huang2020multi}  use a pre-trained single-organ model to generate pseudo labels for un-annotated organs in each dataset, and then constructs a pseudo multi-organ dataset which is used to train the multi-oragn segmentation model. Since the quality of pseudo labels is essential, a penalty that regularizes the organ size distribution of pseudo labels is introduced in \cite{zhou2019prior}, and a pair of weight-averaged model is proposed in \cite{huang2020multi} in which each model is trained with the outputs of the other model to get more reliable pseudo labels. Most of the existing pseudo label based methods only utilize the binary outputs of the multiple pre-trained single-organ models. Learning from soft pseudo labels via consistency loss has been proved to be more effective than learning from hard labels \cite{zou2020pseudoseg}.

\begin{figure}[t]
\center
\includegraphics[width=0.9\textwidth]{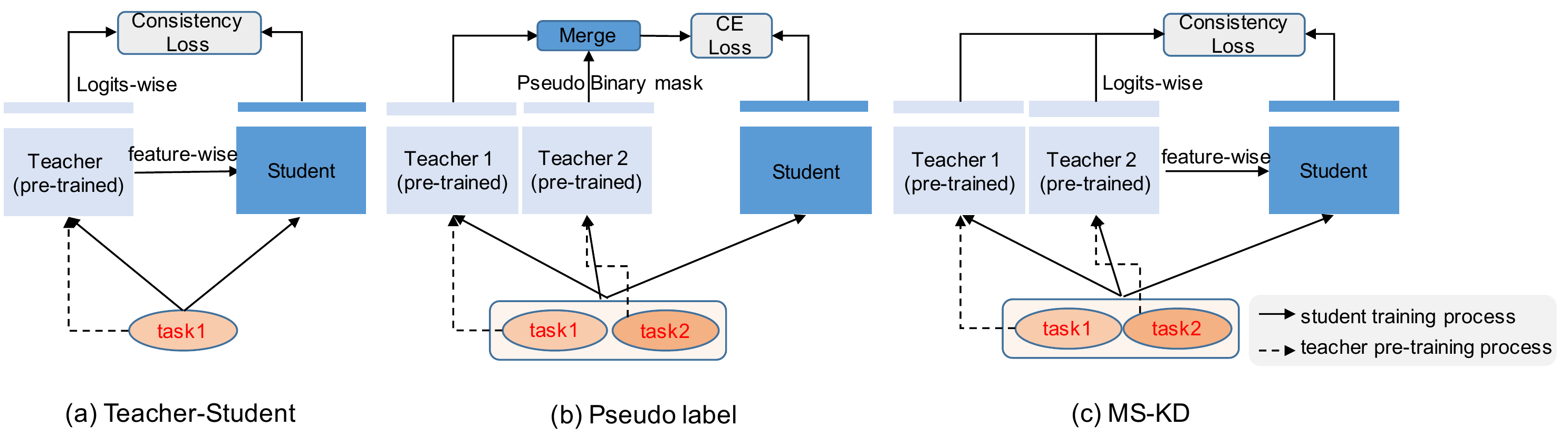}
\caption{(a) Typical knowledge distillation framework (b) Pseudo label based method (c) MS-KD: Multi-teacher Single-student Knowledge Distillation framework.}
\label{intro}
\end{figure}

In this paper, we attempt to explore a knowledge distillation framework for learning to segment multiple organs combining a set of single-organ datasets.  
In particular, we employ teacher-student models \cite{hinton2015distilling}, widely used for knowledge distillation for image classification \cite{urban2016deep} and segmentation \cite{xie2018improving} tasks, which utilizes the soft pseudo labels or intermediate features from the teacher model \cite{zagoruyko2016paying,liu2019structured} to train the student model.
In our case, we have multiple teacher models, each of which is a single-organ segmentation model. It should be noted that the above setting is different from a typical teacher-student model not only in the number of teachers, but also in that each of the teacher networks is for a different single organ segmentation task and they jointly teach the student to segment multiple organs (Fig \ref{intro}).
The classical consistency learning used by teacher-student framework thus can not be directly applied. 

To tackle the above challenge, 
We propose a Multi-teacher Single-student Knowledge Distillation (MS-KD) framework, to distillate the teacher models into a single student model for multi-organ segmentation.
A region-based supervision is proposed to enable the student to learn each task from the respective teacher.
The student learns segmenting an organ from the respective teacher and learn segmenting background from all the teachers, so each teacher supervises the student in two regions, the organ region predicted by the teacher and the background region where all teachers agree. Specifically, logits-wise and feature-wise supervision are employed for each region. As the logits of the teacher networks and the student network are of different dimention, \emph{i.e.,} two vs. $K+1$ where $K$ is the number of organs, for logits-wise supervision, a simple dimension transfer is applied to the logits outputs of teacher, and the transferred outputs are used to constrain the student. For feature-wise supervision, the distribution of the student intermediate features is constrained according to the teachers intermediate features.
We evaluate the proposed method in four public datasets, including three single-organ datasets for both training and test, and one multi-organ datasets only for test. Experimental results shows that the MS-KD has the capacity to enable student to learn from and surpass the multiple teachers.

The contributions of this paper are summarized as follows. (1) We propose a novel Multi-Teacher Single-Student Knowledge Distillation (MS-KD) framework to train a multi-organ segmentation network from multiple single-organ segmentation networks. (2) Region-based supervision methods, including logits-wise supervision and feature-wise supervision, are proposed to enable student model to learn from multiple task-different teachers.

\section{Methodology}
\begin{figure}[t]
\includegraphics[width=\textwidth]{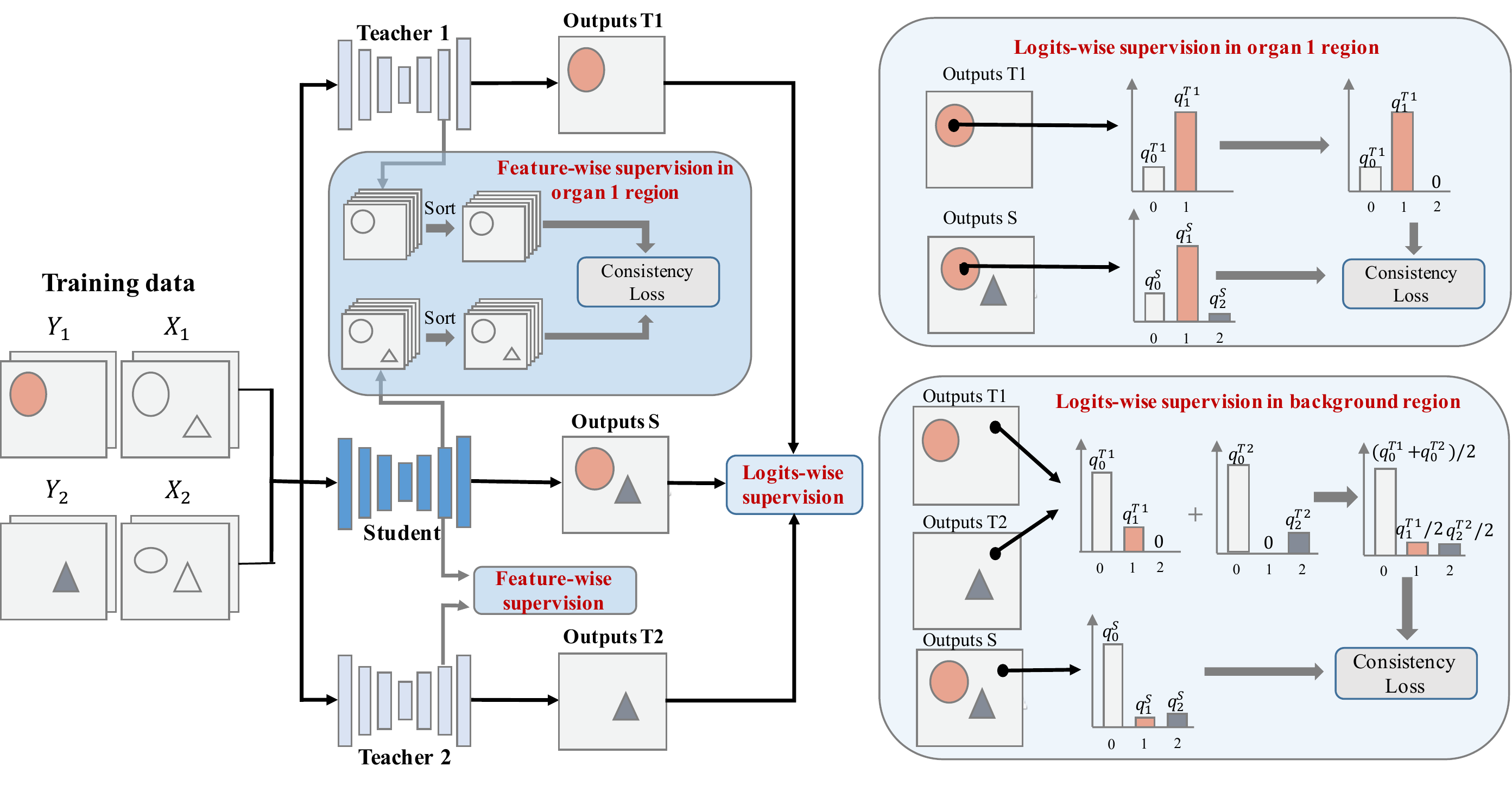}
\caption{Overview of MS-KD framework. The framework uses the union of images from the $K$ single-organ datasets ($K=2$ here). The framework consists of $K$ pre-trained single-organ segmentation networks as teacher models, and a multi-organ segmentation network as student model. The student is trained using region-based logits-wise supervision and feature-wise supervision.}
\label{architecture}
\end{figure}
Given a set of single organ datasets $\{D_1, ..., D_K\}$, where the dataset for the $k$-th organ $D_k$ contains a set of input images $X^k=\{x_i^k\}_{i=1}^{N_k}$ and the corresponding set of binary segmentation masks $Y^{k}=\{y_i^k\}_{i=1}^{N_k}$. The label of each pixel is binary, \emph{i.e.,} 0 for background and  1 for organ-$k$. 
The goal is to train a $K$-organ segmentation model with the above single organ datasets.

\subsection{Multi-Teacher Single-Student Knowledge Distillation (MS-KD)}
The overall framework of the MS-KD framework is illustrated in Fig \ref{architecture},  consisting of $K$ single-organ segmentation model as the teachers and one multi-organ segmentation model as the student.

The  $K$ single-organ segmentation teacher networks $\{f_{T_1}, f_{T_2},...,f_{T_K}\}$ are trained on the corresponding single organ datasets, each with the loss function
\begin{equation}
\label{eq1}
\min_{\theta_k}\sum_{i=1}^{N_k}\mathcal{L}(f_k(x_i^k;\theta_k),y_i^k),
\end{equation}
where $\theta_k$ denotes the network parameters of teacher-$k$ and the loss function $\mathcal{L}$  is the combination of dice loss and cross-entropy loss. The teachers are frozen when training the student. The student has the same architecture as the teachers, except for the dimension of outputs (2 vs. $K+1$). Different from existing Teacher-student models, not only the proposed framework has more than one teachers, but also the task of student is different from the task of any teacher, which means the classical consistency learning method are not directly applicable. 

\subsection{Region-based Supervision}
Here each teacher model is specialized in segmenting one individual organ. 
For an image $x$, denote $M^{k}(x)$ as its binary predictions from the teacher-$k$:
\begin{equation}
\label{equ_Mask}
M^{k}(x_i)=\left\{
\begin{aligned}
&1, \text{pixel $i$ with organ-k predictions} \\
&0, \text{pixel $i$ with background predictions}.
\end{aligned}
\right.
\end{equation}
As each teacher model predicts the corresponding task-irrelevant organs as background, the background predictions of the teacher-$k$ is actually the union of true background region and $K-1$ irrelevant organ regions. So for each teacher, we can only trust its prediction of the corresponding organ regions.  Considering that the organs are expected to occupy mutually exclusive regions, a region-based supervision strategy is proposed to tackle this challenge. Before we go to the details, we first define two sets of regions.
\begin{itemize}
    \item \textbf{Organ-$k$ region}: the pixels $\{x_i | \forall x_i \in x,  M^{k}(x_i)=1$\}. By supervising in this region, student learns to segment organ-$k$ from the teacher-$k$.
    \item \textbf{True background region}:  The true background region is thus obtained by intersecting the background regions of all teachers. Let $M^B = (1-M^{1}) \cap (1-M^{2})...\cap (1-M^{K})$. The background region is thus 
    the pixels $\{x_i | \forall x_i \in x,  M^{B}(x_i)=1$\}.
\end{itemize}
The set of organ-$k$ regions represent the specific regions that the teacher-$k$ is an expert, while the true background region represents where the collective intelligence from all teachers are located. The dentition of the above regions enable us to limit the teacher's influence on the student to the regions that the teachers are more likely to be right. The student model then learns to segment multiple organs combing supervisions from set of non-overlapping specialized teachers. 

\subsubsection{Region-based logits-wise supervision}

The dimensions of the teachers' logits outputs of   are different from that of the student, \emph{i.e.,} two vs. $K+1$. To deal with such situations, we propose a simple but effective logits dimension transferring method.
Suppose $q_c^{k}$ is the classification probability of class $c$ of the teacher-$k$, computed by softmax function  $q_{c}^{k}=\frac{f_k(x_i^k;\theta_k)_c}{\sum_c exp( f_k(x_i^k;\theta_k)_c)}$, where $c \in \{0, 1\}$.
To transfer $q^{k} \in R^2$ to the same dimension as the student's logits, 
we map the supervision signal from the teacher-$k$ to $\hat{q}^{k} \in R^K$ as follows  
\begin{equation}
\label{eq2}
\hat{q}_c^{k}=\left\{
\begin{aligned}
&q_0^{k}, & \text{if}~c=0\\
&q_1^{k}, & \text{if}~c=k\\
&0, & \text{if}~c \not \in \{0, k\}
\end{aligned}
\right. \ \ \ \forall c \in \{0, ..., K\}.
\end{equation}
In this way, the supervision signal for the organ-$k$ region is generated only from the outputs of the teacher-$k$.
The student's logits outputs in the organ-$k$ region is constrained with $\hat{q}_c^{k}$ in region $M_i^k=1$ . The logits-wise loss for the organ-$k$ is thus defined as
\begin{equation}
\label{eq3}
L_{\text{logit}}^{k} = \frac{1}{W\times H}\sum_{i\in \mathcal{R}} KL(\hat{q}_i^{k}, q_i^S) \cdot M^{k}_i,
\end{equation}
where $i$ denotes the $i$th pixel, $\mathcal{R}=\{1,2,...,W\times H\}$ denotes all the pixels, $q^S$ is the classification probability of the student, and $KL(\cdot)$ is the Kullback-Leibler divergence.
For the true background region, the supervision signal is generated from classification probabilities of all teachers,
\begin{equation}
\label{eq4}
\hat{q}^B=(\hat{q}^{1}+\hat{q}^{2}+...+\hat{q}^{K})/K.
\end{equation}
The logits-wise loss for background region $M_i^B=1$ can be formulated as 
\begin{equation}
\label{eq5}
L_{\text{logit}}^{B} = \frac{1}{W\times H}\sum_{i\in \mathcal{R}}KL(\hat{q}_i^B, q_i^S) \cdot M^{B}_i.
\end{equation}

\subsubsection{Region-based feature-wise supervision}
For region-based supervision in each intermediate layer,
we denote $M^{k,l}$ as the binary mask for intermediate layer of teacher-$k$, generated by down-sampling $M^{k}$ for $l$ times. We hypothesis that in organ-$k$ region ($M^{k,l}=1)$ of features, features distribution of teacher-$k$ and that of student should be similar. Since the teachers are trained separately, the permutations of features are different within these teachers, thus we can not directly constrain features between teachers and student. We sort the features for each pixel along the channel dimension and then supervise in the organ regions only. For the layer corresponding to $M^{k,l}$, the feature-wise loss is computed as,
\begin{equation}
\label{eq6}
L_{\text{feature}}^{k,l} = \frac{1}{(W/2^l)\times (H/2^l) }\sum_{i\in \mathcal{R}^l}KL(\text{sort}(F^{k,l}_i), \text{sort}(F^{S,l}_i)) \cdot M^{k,l}_i,
\end{equation}
where $F^{k,l}$ and $F^{S,l}$ are the features after softmax of the teacher-$k$ and the student, respectively, and $\mathcal{R}^l=\{1,2,...,(W/2^l)\times (H/2^l)\}$.

\subsubsection{Overall segmentation loss}
The total loss is formulated as follows,
\begin{equation}
\label{eq7}
L = \sum_{k=1}^KL_{\text{logit}}^{k}+\lambda_1 L_{\text{logit}}^{B}+\lambda_2\sum_{k=1}^KL_{\text{feature}}^{k,l},
\end{equation}
where $\lambda_1$ and $\lambda_2$ are the respective weights of the loss.

\section{Experiments}
\subsection{Datasets and impelmentation details}
To evaluate the proposed framework, our work utilizes three single-organ datasets, MSD-Spleen \cite{simpson2019large} with spleen annotations, KiTS \cite{heller2019kits19} with kidney annotations, NIH-Pancreas \cite{roth2016data} with pancreas annotations, and a multi-organ dataset, BTCV.
BTCV  is a multi-organ dataset with 30 scans. It provides segmentation masks for all abdominal organs except the duodenum. All single-organ datasets are divided into training and test sets with 4:1 ratio and the BTCV dataset is used only for test. We use Dice-Score-Coefficient (DSC) and Hausdorff Distance (HD) as the evaluation metrics. 

For data preprocessing, we truncted the HU values in each scan to the range $[-325, 325]$ to have a better contrast on the abdominal organs and linearly normalized to $[-1, +1]$. We choose 2D-UNet as our segmentation backbone and input patch size is $512\times 512$. During training, we use batch size 4 and train 250 iterations per epoch. The maximum epoch is 1000. The proportion of patches that contain foreground in each batch is set to be 33\%. Adam is used for optimizer and the initial learning rate of the network is 3e-4. When the loss reduction is less than 1e-3, the learning rate decays by 20\%. 
We empirically choose the features from decoder and $l=1$, and the hyper-parameter $\lambda_1$ is 1, $\lambda_2$ is 10. 
To ensure a fair comparison, the same training strategies are applied to all competing methods. 

\subsection{Ablation study}
\begin{table}[t]
\caption{Performance comparison between MS-KD and competing methods in single-organ datasets. 
The best result for each column is marked in bold.}
\label{ablation}
\center
\begin{tabular}{c|cccc|cccc}
\hline
&  \multicolumn{4}{c}{DSC(\%)} & \multicolumn{4}{|c}{HD(mm)}\\
\hline
Method &  Spleen & Kidney & Pancreas & Avg & Spleen & Kidney & Pancreas & Avg\\
\hline
Individual & 95.86 & 96.06 & 78.99 & 90.30 & 9.90 & 14.39 & 18.43  & 14.24\\
LW &  95.97 & \pmb{96.55} & 80.76 & 91.10 & 9.86 & \pmb{12.42} & 17.12  & 13.13\\
MS-KD (LW+FW) & \pmb{96.16} & 96.21 & \pmb{81.59} & \pmb{91.32} & \pmb{8.81} & 13.43 & \pmb{16.85} & \pmb{13.03}\\
\hline
Co-training \cite{huang2020multi}& 96.11 & 95.81 & 80.76 & 90.89 & 10.57 & 13.09 & 18.33 & 14.00  \\
DoDNet \cite{zhang2020dodnet} & 96.00 & 96.04 & 79.11 & 90.42 & 9.84 & 13.07 & 19.44 & 14.12\\
\hline
\end{tabular}
\end{table}

We first train three single-organ segmentation models separately for each organ, denoted as ``Individual" in Table \ref{ablation}. 
It achieves $95.86\%$, $96.06\%$, $78.99\%$ DSC for spleen, kidney, and pancreas, respectively, and an average DSC of 90.30\%. 
We then use the individual models as teachers to train the student model. After applying the region-based logits-wise supervision, denoted as ``LW", the MS-KD framework gets a remarkable improvement of $0.80\%$ in terms of average DSC ($91.10\%$ vs $90.30\%$). Specifically, the average performance improvement is mainly obtained by the more challenging organ, pancreas ($80.76\%$ vs $78.99\%$). And the other two organs get slightly better performance compared to individual model, which demonstrates the effectiveness of the region-based logits-wise supervision to enable student model to learn from all the teachers. The region-based feature-wise supervision, denoted as ``FW", further boosts the performance of MS-KD framework. The average DSC increases by $0.22\%$ ($91.32\%$ vs $91.10\%$). We also conduct experiments on BTCV dataset to evaluate the generalization ability of MS-KD framework, shown in Table \ref{btcv}. The model is trained on single-organ datasets and test on BTCV dataset. MS-KD gets a remarkable improvement 1.32\% for average DSC compared to individual models. It demonstrates that training with union of binary-labeled datasets can obtain much better performance on generalization.

\begin{table}[tb]
\caption{Performance comparison between MS-KD and competing methods in BTCV dataset. The best result for each column is marked in bold.}
\label{btcv}
\center
\begin{tabular}{c|cccc|cccc}
\hline
&  \multicolumn{4}{c}{DSC(\%)} & \multicolumn{4}{|c}{HD(mm)}\\
\hline
Method &  Spleen & Kidney & Pancreas & Avg & Spleen & Kidney & Pancreas & Avg\\
\hline
Individual & 91.90 & 88.11 & 70.51 & 83.51 & 10.66& 20.85 & 22.43&17.98\\
LW         & 92.29 & \pmb{88.78} & 72.64 & 84.57 & 10.05 & 17.87 & \pmb{19.20} & 15.71\\
MS-KD (LW+FW)  & \pmb{92.43} & 88.77 & \pmb{73.29} & \pmb{84.83} & \pmb{9.61} & \pmb{17.52} & 19.32 & \pmb{15.48}\\
\hline
Co-training \cite{huang2020multi} & 92.10 & 88.57 & 72.53 & 84.40 & 10.29 & 18.58 & 19.30 & 16.06 \\
DoDNet \cite{zhang2020dodnet} & 91.93 & 88.51 & 71.67 & 84.03 & 10.61 & 19.11&21.72 & 17.14\\

\hline
\end{tabular}
\end{table}

\begin{figure}[t]
\begin{center}
\includegraphics[width=0.8\linewidth]{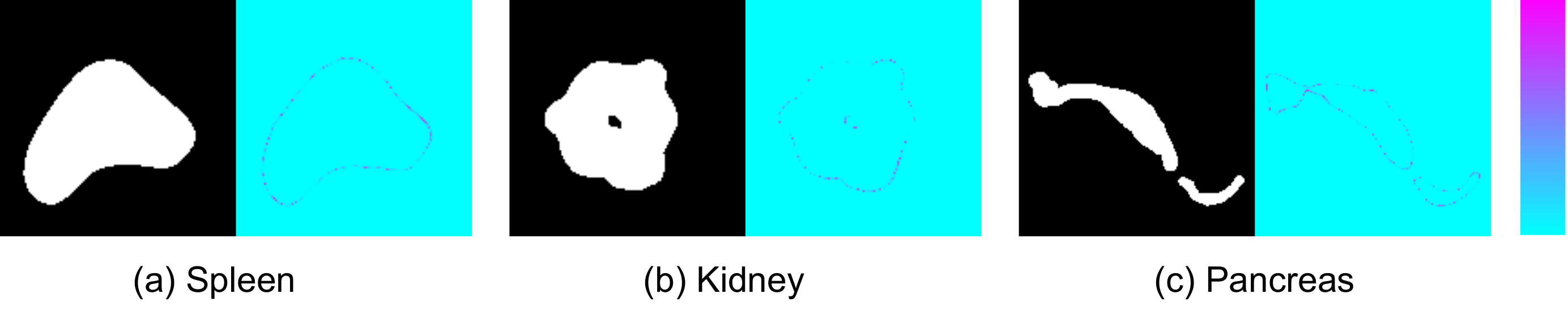}
\end{center}
\vspace{-0.5cm}   \caption{Ground truth and uncertainty region of single-organ teacher. Darker color means the teacher is more uncertain in the region.}
\label{logits}
\end{figure}

\subsection{Comparison with state-of-the-art methods}
\begin{figure}[tb]
\center
\includegraphics[width=\linewidth]{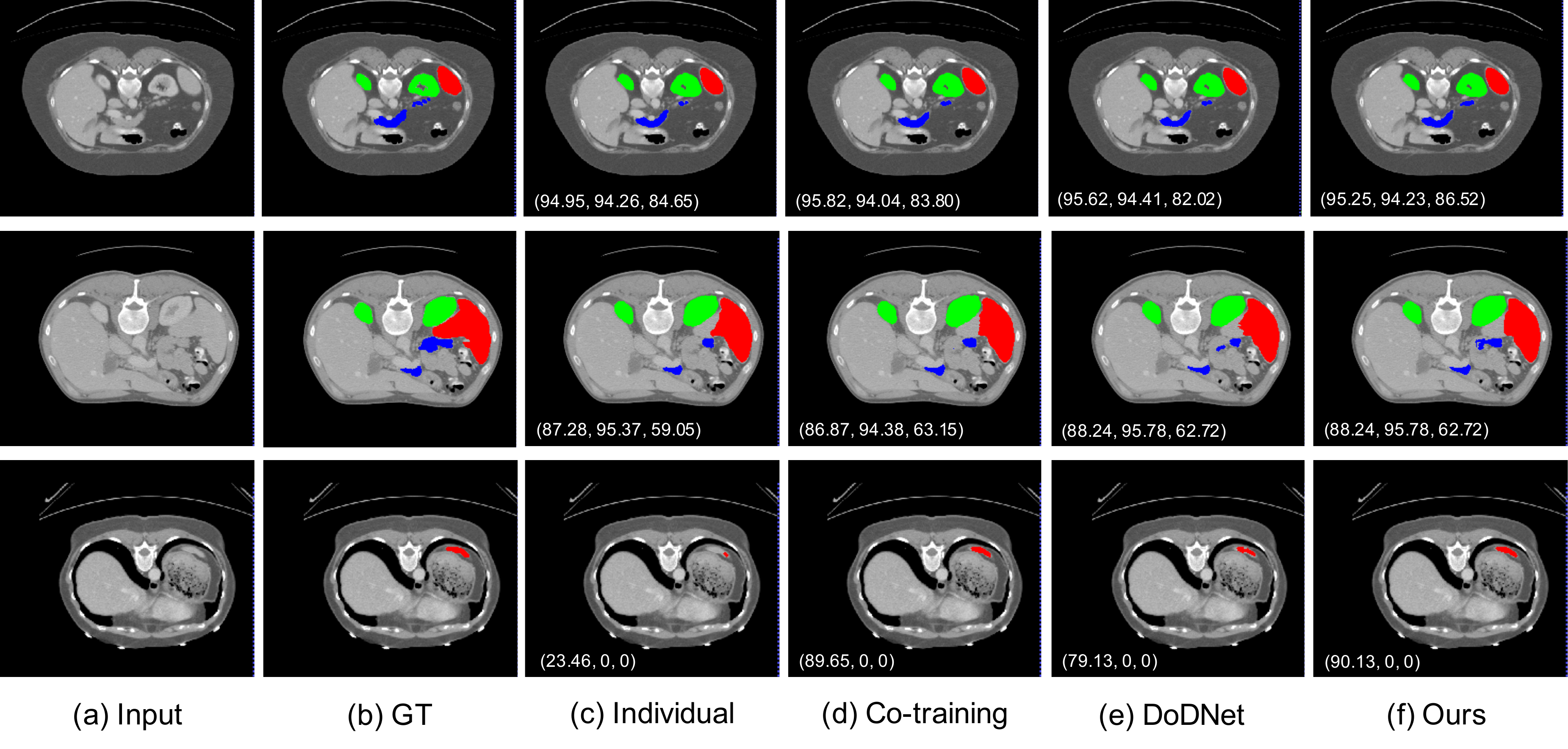}
\center
\vspace{-0.5cm}
\caption{Visualization of segmentation results obtained by different methods. The organs are annotated in color: red for spleen, green for kidney, blue for pancreas. The DSC for each slice is shown in parenthesis.}
\label{visualize}
\end{figure}

We also compare the proposed framework with two recently-developed partially-supervised segmentation networks. The Co-training refers to the work by Huang et al. \cite{huang2020multi} which collaboratively trains a pair of networks and let the two networks generate pseudo labels to teach each other. The DoDNet refers to the work by Zhang et al. \cite{zhang2020dodnet} which generates dynamic head conditioned on both task and input image. It shows that (1) Co-training achieves better performance than individual networks, which demonstrates that using more partially labeled training data can increase the segmentation performance; (2) pseudo-label based method performs better than condition-based method. We suspect that it is because the former utilized the unlabeled organs in other datasets for training via generated pseudo labels while the latter not. (3) MS-KD outperforms these two methods by a considerate gain 0.43\% (91.32\% vs 90.89\%).
As shown in Fig \ref{logits}, the most uncertain region is the edge of the organ. Using smooth supervision signal produced by logits outputs of teachers other than using binary pseudo labels, can better utilize the pre-trained models.
In addition, it should be noticed that the Co-training method needs to train a pair of networks simultaneously, which means its parameters number is nearly twice as much as ours, and the DoDNet needs to generate segmentation head dynamically and inference $K$ times to get the final segmentation results. The proposed MS-KD framework gets better performance without any additional computation cost for both training and inference. To make a qualitative comparison on all the three organs, we  visualized the segmentation results obtained by four methods on BTCV datasets in Fig \ref{visualize}.

\section{Conclusion}
We propose Multiple-Teacher Single-Student Knowledge Distillation (MS-KD) framework for multi-organ segmentation with several single-organ datasets. Teacher models are single-organ segmentation models pre-trained with single-organ datasets, respectively. To enable the student learns from multiple task-different teachers, the novel region-based logits-wise supervision and feature-wise supervision methods are developed. We demonstrate the effectiveness of our framework in four public datasets and surpass the recently-developed methods.

\bibliographystyle{splncs04}
\bibliography{egbib}
\end{document}